\documentclass[11pt,a4paper]{article}
\usepackage[hyperref]{acl2020}
\usepackage{times}
\usepackage{latexsym}

\usepackage{tabularx}
\usepackage{booktabs}
\usepackage{multirow}
\usepackage{color}
\usepackage{amssymb}
\usepackage{bm}
\usepackage{graphicx}
\usepackage{algorithm}
\usepackage{algorithmicx}
\usepackage{CJK}
\usepackage{algpseudocode}
\algnewcommand\algorithmicinput{\textbf{Input:}}
\algnewcommand\INPUT{\item[\algorithmicinput]}
\algnewcommand\algorithmicoutput{\textbf{Output:}}
\algnewcommand\OUTPUT{\item[\algorithmicoutput]}

\usepackage{microtype}

\aclfinalcopy 
\def\aclpaperid{2794} 

\newcommand*\myat{{\fontfamily{ptm}\selectfont\small @}}

\title{CDL: Curriculum Dual Learning for \\ Emotion-Controllable Response Generation}

\def\first{$^1$}
\def\second{$^2$}
\def\comma{$^,$}
\def\star{$^*$}

\author{Lei Shen\first\comma\second 
~~~ Yang Feng\first\comma\second\star
\\
{\first {Key Laboratory of Intelligent Information Processing,}} \\ Institute of Computing Technology, Chinese Academy of Sciences, Beijing, China  \\
{\second {University of Chinese Academy of Sciences, Beijing, China}} \\
{\tt \{\href{mailto:shenlei17z@ict.ac.cn}{shenlei17z}, \href{mailto:fengyang@ict.ac.cn}{fengyang}\}\myat ict.ac.cn}	
}

\date{}

\begin{document}
\maketitle
\newcommand\blfootnote[1]{%
\begingroup 
\renewcommand\thefootnote{}\footnote{#1}%
\addtocounter{footnote}{-1}%
\endgroup
}
\blfootnote{\star{Yang Feng is the corresponding author.}}
\begin{abstract}
Emotion-controllable response generation is an attractive and valuable task that aims to make open-domain conversations more empathetic and engaging. Existing methods mainly enhance the emotion expression by adding regularization terms to standard cross-entropy loss and thus influence the training process. However, due to the lack of further consideration of content consistency, the common problem of response generation tasks, safe response, is intensified. Besides, query emotions that can help model the relationship between query and response are simply ignored in previous models, which would further hurt the coherence. To alleviate these problems, we propose a novel framework named Curriculum Dual Learning (CDL) which extends the emotion-controllable response generation to a dual task to generate emotional responses and emotional queries alternatively. CDL utilizes two rewards focusing on emotion and content to improve the duality. Additionally, it applies curriculum learning to gradually generate high-quality responses based on the difficulties of expressing various emotions. Experimental results show that CDL significantly outperforms the baselines in terms of coherence, diversity, and relation to emotion factors.
\end{abstract}

\section{Introduction}

Infusing emotions into dialogue systems can make conversational agents more human-like and benefit the interaction between human and machine \cite{prendinger2005empathic,prendinger2005using,partala2004effects}. In some real-life scenarios, we need to customize and control the agent's emotion so that the agent can express a specific one. For example, in psychological counseling, the agent is supposed to express sadness to show the sympathy and also convey happiness to cheer the patient up.

\begin{table}[htb]
\small
\centering
\begin{tabular}{l|lp{5.8cm}}
  \toprule[1pt]
  \multirow{8}*{1} & $q$ & It is very pleasant to have a cup of black tea with sugar on a cold day. (Happy)\\ \cline{2-3}
  & $r_1$ & [Neural] It starts to cool down today. \\ \cline{2-3}
  & $r_2$ & [\textcolor[rgb]{1,0.08,0.58}{Like}] I will try, thanks for your advice.  \\ \cline{2-3}
  & $r_3$ & [\textcolor[rgb]{0.42,0.35,0.80}{Sad}] I am frozen to death ... \\ \cline{2-3}
  & $r_4$ & [\textcolor[rgb]{0,0,0.7}{Disgust}] Winner is the worst season. \\ \cline{2-3}
  & $r_5$ & [\textcolor[rgb]{0.44,0.50,0.56}{Angry}] You know nothing! \\ \cline{2-3}
  & $r_6$ & [\textcolor[rgb]{0.7,0,0.2}{Happy}] I really like to drink black tea. \\ 
  \midrule[1pt]
  \multirow{3}*{2} & $q$ & So pets live better than humans now... (Sad) \\ \cline{2-3}
  & $r_1$ & [\textcolor[rgb]{0,0,0.7}{Disgust}] You are so bad. \\ \cline{2-3}
  & $r_2$ & [\textcolor[rgb]{0.7,0,0.2}{Happy}] Haha, you too. \\ 
  \midrule[1pt]
  \multirow{2}*{3} & $q$ & We should study hard. (Neural) \\ \cline{2-3}
  & $r$ & [\textcolor[rgb]{0,0,0.7}{Disgust}] You are so bad. \\ 
  \midrule[1pt]
  \multirow{4}*{4} & $q$ & Happy birthday, Xinxin. May you be more beautiful, find a good person and get married soon! (Happy) \\ \cline{2-3}
  & $r$ & [\textcolor[rgb]{0.7,0,0.2}{Happy}] Haha, you too. \\ 
  \bottomrule[1pt]
\end{tabular}
\caption{Examples of emotion-controllable response generation (response emotions are denoted in brackets). Example 1 is one query and 6 emotional responses. Example 2 and 3 have different queries, but the responses generated with emotion ``Disgust'' are the same. Similar to Example 2 and 4 with emotion ``Happy''. The emotions of queries are marked in parentheses.}
\label{tab:case_for_intro}
\end{table}

Recently, a framework called emotional chatting machine (ECM) \cite{zhou2018emotional} was proposed to address the emotion factor in a controlled manner, which focuses on generating a response with a specific emotion (Example 1 in Table \ref{tab:case_for_intro}). In the research field of emotion-controllable response generation, ECM and its successive methods \cite{colombo2019affect,song2019generating} mainly represent the given emotion category as a vector and add it to the decoding steps to influence the procedure of response generation, which would aggravate the safe response problem. For the response generation task, safe response is notorious, as the model tends to produce some generic but meaningless responses, like ``Thank you'', ``I don't know'', ``Yes'', etc. Due to the constraint of emotion factors, the scale of proper responses shrinks, and the model is more likely to map any query to a frequently-occurring response in that emotion category. That is, given ``Disgust'', the response would be ``You are so bad'' in general, while given ``Happy'', it would be ``Haha, you too'' (Example 2 to 4 in Table \ref{tab:case_for_intro}). 

Intuitively, for a good pair of query and response, they should be in a tight relationship and have equal qualities. Then, both the query-to-response mapping and response-to-query mapping would be easier and more natural. On the contrary, it is hard for a safe response to reach the original query through back-generation, neither on the content level nor the emotion level. 
At the same time, the difficulties of producing various emotions are different, especially in a noisy and uneven-quality dataset. Therefore, we can evaluate the response based on the feedback from the backward process to improve the coherence \cite{zhang2018reinforcing,cui2019dal,luo2019dual} and try to learn from easy to hard data to generate appropriate and emotion-rich responses.

In this paper, we propose a new framework for emotion-controllable response generation named Curriculum Dual Learning (CDL). 
We take the learning of response and query generation with emotions as a dual task, and use the duality to model the mutual relation between them.
The forward and backward models are trained alternatively via reinforcement learning (RL). Rewards designed here aim to encourage both emotion expression and content consistency. Specifically, emotion expression can be either explicit (embodied in some obvious emotion words) or implicit (reflected by the organization of the entire sentence). For example, ``I am happy to meet her again'' is explicit with the word ``happy'', while ``It seems like I have eaten the honey'' is implicit, but the happiness can be felt when we consider the sentence as a whole. Based on these features, we use the accuracy of emotion classification of sentences and the proportion of emotion words as feedbacks for explicit and implicit emotions, respectively. For content consistency, we apply the reconstruction probability as the measurement of coherence (Section \ref{sec:dl}). 
Furthermore, in order to better utilize samples of multiple emotions from the noisy and uneven-quality dataset, we incorporate the curriculum learning (Section \ref{sec:curriculum}) into our dual learning framework (Section \ref{sec:trainingCDL}). 

Experimental results on both automatic and human evaluations show that for a given query and an emotion category, our CDL can successfully express desired emotion as well as keep the response informative and coherent to the query.


\section{Background}
\label{sec:background}
For emotion-controllable response generation, given a query $\bm{q}$ and an emotion category $e_r$, the goal is to generate a response $\bm{r'}$ that is not only meaningful, but also in accordance with the desired emotion. 

{\em Emotional Chatting Machine (ECM)} \cite{zhou2018emotional} addresses the emotion factor using three new mechanisms: Emotion Category Embedding, Internal Memory, and External Memory. Specifically, 1) Emotion Category Embedding models the high-level abstraction of emotion expression by embedding emotion categories, and concatenates corresponding embedding to the input at each decoding step. 2) Internal Memory captures the change of implicit internal emotion states with read and write gates, 3) External Memory applies an external emotion vocabulary to express emotion explicitly, and finally assigns different generation probabilities to emotion and generic words. The loss function on one training sample ($\bm{q}, \bm{
r}$) ($\bm{q}= q_1, q_2, ... , q_n, \bm{r} = r_1, r_2, ... , r_m$) is defined as:
\begin{equation} \label{eq:ecmloss}
    - \sum_{t=1}^m \bm{p}_t log(\bm{o}_t) - \sum_{t=1}^m q_t log(\alpha_t) + || \bm{M}_{e,m}^I||,
\end{equation}
where $\bm{o}_t$ and $\bm{p}_t$ are the predicted token distribution and gold distribution, $\alpha_t$ is the probability of choosing an emotion word or a generic word, $q_t\in \{0,1\}$ is the true choice between them in $\bm{r}$, and $\bm{M}_{e,m}^I$ is the internal emotion state at the last step $m$. The first term is the cross-entropy loss, the second one is used to supervise the probability of selecting an emotion or generic word, and the last one is used to ensure that the internal emotion state has been expressed completely once the generation is finished. Please refer to the original paper for more details.

\section{CDL for Emotion-Controllable Response Generation}
\label{sec:approach}
Since our CDL method is a combination of dual learning (DL) and curriculum learning (CL), we first present the main components of DL, including states, actions, policy and reward, then introduce the plausibility of curriculum learning. Finally, we describe the training algorithm of CDL.

\begin{figure}[htb]
\begin{center}
   \includegraphics[width=0.8\linewidth]{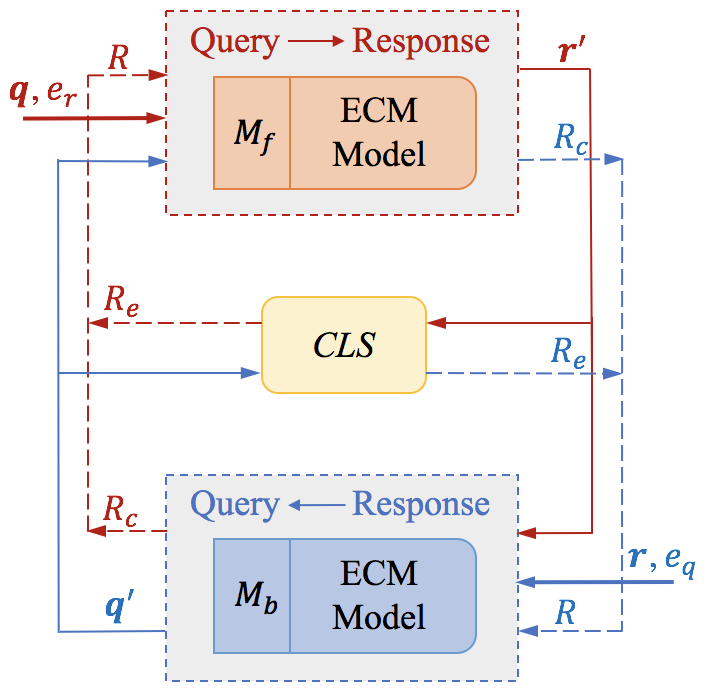}
\end{center}
   \caption{The architecture of dual learning. \textit{CLS}, $M_f$ and $M_b$ are emotion classifier, forward model and backward model, respectively. Red parts are for the forward process, while blue parts are for the backward process.}
\label{fig:overview}
\end{figure}

\subsection{DL Architecture}
\label{sec:dl}
The architecture of DL is illustrated in Figure \ref{fig:overview}. Both the forward model $M_f$ and the backward model $M_b$ are ECMs with independent parameters and are initialized according to the maximum likelihood estimation (MLE). \textit{CLS} is a pre-trained classifier that calculates the score of implicit emotion expression. 

In general, $M_f$ generates a response $\bm{r'}$ for a given query $\bm{q}$ and emotion category $e_r$, and then obtains the reward $R$ that consists of $R_e$ from \textit{CLS} and $R_c$ from $M_b$ (red parts in Figure \ref{fig:overview}). Similarly, $M_b$ generates a query $\bm{q'}$ for a given response $\bm{r}$ and emotion category $e_q$, and obtains the reward $R$ that consists of $R_e$ and $R_c$ from \textit{CLS} and $M_f$ (blue parts in Figure \ref{fig:overview}). These two models are trained alternatively via reinforcement learning (RL). Specifically, an action is the dialogue response to generate. The action space is infinite since arbitrary-length sequences can be generated. A state is denoted by the query, which is further transformed to a vector representation by the encoder. A policy takes the form of a GRU encoder-decoder and is defined by its parameters. Following the work of \citet{li2016deep,zhang2018reinforcing}, we use a stochastic representation of the policy, i.e., a probability distribution over actions given states. 

In order to encourage both content consistency and emotion expression, we introduce two rewards and use them to train $M_f$ and $M_b$. The definition of the two rewards for model $M_f$ is introduced as follows\footnote{Rewards for model $M_b$ can be computed in a similar way, where $q'$, $r$, $b$ and $f$ replace $r'$, $q$, $f$ and $b$, respectively. Therefore, we omit them here for space limitation and brevity.}.

\textbf{Reward for emotion expression}
For implicit emotion expression, a straightforward method is to employ the pre-trained classifier \textit{CLS} to evaluate the emotion category of the generated response $\bm{r'}$, and use the classification accuracy as the reward:
\begin{equation}
    R_{e_1(\bm{q}, \bm{r'})} = p(e_r|\bm{r'}; \varphi),
\end{equation}
where $\varphi$ is the parameter of \textit{CLS}, and it is fixed during training.
For explicit emotion expression, the reward is formulated as:
\begin{equation}
    R_{e_2(\bm{q}, \bm{r'})} = n(w_{e_r}) / |\bm{r'}|,
\end{equation}
where $n(w_{e_r})$ is the number of emotion words belong to category $e_r$, and $|\bm{r'}|$ is the length of $\bm{r'}$.
Then, the emotion reward is defined as:
\begin{equation} \label{eq:emotionreward}
    R_{e(\bm{q}, \bm{r'})} = R_{e_1(\bm{q}, \bm{r'})} + \lambda R_{e_2(\bm{q}, \bm{r'})},
\end{equation}
where $\lambda$ controls the relative importance of implicit and explicit rewards.

\textbf{Reward for content consistency}
If the response are coherent and related to the query, it will be easier to reproduce the query via back generation. Inspired by \citet{zhang2018reinforcing,cui2019dal,luo2019dual}, we measure the coherence by means of reconstructing $q$ conditioned on $r'$. Formally, the content consistency reward is defined as:
\begin{equation}
    R_{c(\bm{q}, \bm{r'})} = p(\bm{q}|\bm{r'}, e_q; \eta),
\end{equation}
where $\eta$ is the parameter of backward model $M_b$, and it is fixed during the training of $M_f$.

\textbf{Overall reward} 
We use the weighted sum of the above two rewards as the final reward:
\begin{equation} \label{eq:overallreward}
    R_{(\bm{q}, \bm{r'})} = R_{c(\bm{q}, \bm{r'})} + \gamma R_{e(\bm{q}, \bm{r'})},
\end{equation}
where $\gamma$ is a hyper-parameter that controls the trade-off between $R_{c(\bm{q}, \bm{r'})}$ and $R_{e(\bm{q}, \bm{r'})}$.

\subsection{Curriculum Plausibility}
\label{sec:curriculum}
Intuitively, learning from less noisy and even-quality dataset is simpler, but in this task, the data is inherently complicated as there are multiple emotions mixed in it. To better utilize the data, we integrate curriculum learning into the dual learning framework. The core of curriculum learning \cite{bengio2009curriculum} is to design an evaluation for complexity, and to provide the model with \textit{easy} samples first, then gradually increase the difficulty. The curriculum is arranged by sorting each sample in training set according to a specific ranking standard. 

Here, We reorder samples from easy, i.e., with high accuracy of emotion classification, to hard. We consider the classification accuracy after pre-training as an indicator of the learning order. Another intuitive way is to put emotionless samples (labelled as ``Neural'') first and then emotional ones, however, it exhibits poor performance in our experiments. 
At training step $t$, a batch of training samples is obtained from the top $f(t)$ portions of the entire sorted training samples. Following \citet{platanios2019competence} and \citet{cai2020learning}, we define the function $f(t)$ as:
\begin{equation} \label{eq:ft}
    f(t) \triangleq \mathrm{min}(1, \sqrt{\frac{t(1-c^2_0)}{T} + c^2_0}),
\end{equation}
where $c^2_0$ is set to 0.01, which means that the model starts training using the 1\%
easiest training samples, and $T$ is a hyper-parameter that represents the duration of curriculum learning (curriculum length). At the early stage of the training process, the  model learns from the samples in the easy part of the curriculum, where there is only one emotion category. As the advance of the curriculum, the difficulty gradually increases, as complex training samples from more different categories appear. After training $T$ batches, training sample of each batch is drawn from the whole training set, which is the same as the conventional training procedure. 

\subsection{Training of CDL}
\label{sec:trainingCDL}

\textbf{Optimization}
We use the policy gradient method \cite{williams1992simple} to find parameters that lead to a larger expected reward. For the forward learning process, the expected reward of the generated response $\bm{r'}$ and its approximate gradient are defined as:
\begin{equation}
    J(\theta) = \mathbb{E}[R_{(\bm{q}, \bm{r'})}],  
\end{equation}
\begin{equation} \label{eq:opt1}
    \nabla_\theta J(\theta) \simeq R'_{(\bm{q}, \bm{r'})} \cdot \nabla_\theta \mathrm{log}(p_\theta(\bm{r'}|\bm{q}, e_r)),
\end{equation}
where $\theta$ is the parameter of forward model $M_f$, $R'_{(\bm{q}, \bm{r'})} = R_{(\bm{q}, \bm{r'})} - b_f$, and $b_f$ is the baseline value from the greedy search decoding method for $M_f$, which is used to reduce the variance of the estimation \cite{zaremba2015reinforcement,paulus2017deep}. 
Analogously, for the backward learning process, the expected reward of the generated query $\bm{q'}$ and corresponding approximate gradient are defined as:
\begin{equation}
    J(\eta) = \mathbb{E}[R_{(\bm{r}, \bm{q'})}],  
\end{equation}
\begin{equation} \label{eq:opt2} 
    \nabla_\eta J(\eta) \simeq R'_{(\bm{r}, \bm{q'})} \cdot \nabla_\eta \mathrm{log}(p_\eta(\bm{q'}|\bm{r}, e_q)),
\end{equation}
where $\eta$ is the parameter of backward model $M_b$, $R'_{(\bm{r}, \bm{q'})} = R_{(\bm{r}, \bm{q'})} - b_b$, and $b_b$ is the baseline value from the greedy search decoding method for $M_b$.

\begin{algorithm}[H] 
\caption{Curriculum dual learning algorithm for emotion-controllable response generation} 
\label{alg:CDL} 
\begin{algorithmic}[1]
\INPUT The training set $\mathcal{D} = \{(q_i, e_{q_i}, r_i, e_{r_i})\}$ where each query-response pair is labelled with corresponding emotion labels $e_{q_i}$ and $e_{r_i}$
\OUTPUT $M_f$ and $M_b$
\State Pre-train $M_f$ and $M_b$ with $(q_i, r_i, e_{r_i})$ and $(r_i, q_i, e_{q_i})$, respectively, based on Eq. \ref{eq:ecmloss}
\State Pre-train \textit{CLS} with $(q_i, e_{q_i})$ and $(r_i, e_{r_i})$
\State Sort training samples according to the ranking standard in Section \ref{sec:curriculum} for both forward and backward learning process to get $\mathcal{D}_f$ and $\mathcal{D}_b$
\For{training step $t=1, ..., T$} 
\State \Comment{Train $M_f$}
\State Sample a batch $B_{f_t}$ in $\mathcal{D}_f$ based on Eq. \ref{eq:ft}
\State Sample $(q,r,e_r)$ from $B_{f_t}$
\State Generate response $r'$ via $M_f$
\State Compute reward $R$ based on Eq. \ref{eq:overallreward}
\State Update $\theta$ using $R$ based on Eq. \ref{eq:opt1}
\State Teacher Forcing: Update $\theta$ with $(q,r,e_r)$
\State \Comment{Train $M_b$}
\State Sample a batch $B_{b_t}$ in $\mathcal{D}_b$ based on Eq. \ref{eq:ft}
\State Sample $(r,q,e_q)$ from $B_{b_t}$
\State Generate response $q'$ via $M_b$
\State Compute reward $R$ based on Eq. \ref{eq:overallreward}
\State Update $\eta$ using $R$ based on Eq. \ref{eq:opt2}
\State Teacher Forcing: Update $\eta$ with $(r,q,e_q)$
\EndFor
\end{algorithmic}
\end{algorithm}

\noindent\textbf{Teacher Forcing} 
When $M_f$ and $M_b$ are trained with only the rewards from the dual tasks, the training process would easily collapse as it may find an unexpected way to achieve a high reward but fail to guarantee the fluency or readability of the generated text \cite{ranzato2015sequence,pasunuru2018multi,luo2019dual}. To stabilize the training process, after each update according to Eq. \ref{eq:opt1} or \ref{eq:opt2}, $M_f$ or $M_b$ is exposed to real query-response pairs and is trained via MLE, which is also known as Teacher Forcing \cite{li2017adversarial,lamb2016professor}. 

The training procedure of CDL is summarized in Algorithm \ref{alg:CDL}. First, we use MLE to pre-train $M_f$, $M_b$ and \textit{CLS} with query-response pairs and emotion labels in the training set. After the pre-training phase, we sort samples in the training set following the ranking standard in Section \ref{sec:curriculum}. For forward learning process, the ranking is based on responses, while for backward learning process, it is based on queries. Then, we can get two sorted training set $\mathcal{D}_f$ and $\mathcal{D}_b$ for each direction. Finally, $M_f$ and $M_b$ are optimized with rewards and the regularization of Teacher Forcing, alternatively.

\section{Experiments}
\label{sec:experiments}
In this section, we conduct experiments to evaluate our proposed method. We first introduce some empirical settings, including dataset, hyper-parameters, baselines, and evaluation measures. Then we illustrate our results under both automatic and human evaluations. Finally, we give out some cases generated by different models and do further analyses over our method.

\subsection{Dataset}
We apply our method on the corpus of NLPCC 2017 Emotional Conversation Generation Challenge\footnote{\url{http://coai.cs.tsinghua.edu.cn/hml/challenge2017/}}, namely NLPCC2017 Dataset, which is an extension version of the dataset collected by \citet{zhou2018emotional}. The provided dataset is already segmented into Chinese words. There are over 1 million query-response pairs, in which both the query and response are labelled with one emotion tag among ``Happy'', ``Angry'', ``Disgust'', ``Sad'', ``Like'' and ``Neutral''. The dataset has been tokenized into words. We randomly split the whole dataset into training/validation/test set with the number of 1,105,487/11,720/2,000. The detailed statistics of training set are shown in Table \ref{tab:training_set_statistics}.

\begin{table}[!htb]
    \small
    \centering
    \begin{tabular}{c|lrr} 
    \toprule[1pt]
        \bf \multirow{8}*{Training} & \bf Emotion & \bf Query & \bf Response \\ \hline
        & Happy & 120,358 & 197,528 \\ 
        & Angry & 79,611 & 138,198 \\ 
        & Disgust & 184,427 & 197,428 \\ 
        & Sad & 128,482 & 179,215 \\ 
        & Like & 257,471 & 197,565 \\
        & Neutral & 335,138 & 195,553 \\ \cline{2-4}
        & \multicolumn{3}{c}{1,105,487} \\ \hline
        \bf Validation & \multicolumn{3}{c}{11,720} \\ \hline
        \bf Test & \multicolumn{3}{c}{2,000} \\
    \bottomrule[1pt]
    \end{tabular}
    \caption{Statistics of the NLPCC2017 Dataset. In the training set, we count the number of queries and responses for each emotion category.}
    \label{tab:training_set_statistics}
\end{table}

\subsection{Hyper-parameter Settings}
The settings of both $M_f$ and $M_b$ follow the default implementation details of original ECM paper \cite{zhou2018emotional}, where the encoder and decoder have 2-layer GRU structures with 256 hidden cells for each layer, the embedding size of words and emotion categories are set to 100, and the vocabulary size is limited to 40,000. 
The minimum and maximum sentence length is set to 3 and 30, respectively. We train a TextCNN-based classifier \cite{kim2014convolutional} and the classification accuracy reaches 65.6\% on the test set, which has the similar performance with those used by \cite{zhou2018emotional} and \cite{song2019generating}.
Before curriculum dual learning, model $M_f$ and $M_b$ are pre-trained 10 epochs via MLE. The optimizer is Adam \cite{kingma2014adam} with 0.05 initial learning rate for pre-training and $10^{-5}$ for curriculum dual learning. The batch size is set to 64. $\lambda$ in Eq. \ref{eq:emotionreward} is 0.5, $\gamma$ in Eq. \ref{eq:overallreward} is 1 and $T$ in Eq. \ref{eq:ft} is 100k. During curriculum dual learning, training runs until the performance on validation set does not improve.

\subsection{Baselines}
We compare our approach with four representative baselines: (1) \textbf{S2S-Attn}: The Seq2Seq model with attention mechanism as in \citet{shang2015neural}. (2) \textbf{EmoEmb}: A Seq2Seq variant which takes the embedding of emotion categories as additional input at each decoding position \cite{ficler2017controlling,li2016persona}. (3) \textbf{EmoDS}: An emotional dialogue system with lexicon-based attention and a word-based classifier \cite{song2019generating}. (4) \textbf{ECM}: Emotional Chatting Machine proposed by \citet{zhou2018emotional}.

Additionally, we also conduct ablation study to better analyze our method as follows: (5) \textbf{CDL-emo}: CDL with emotion reward only; (6) \textbf{CDL-con}: CDL with content reward only, which is similar to the work of \citet{zhang2018reinforcing}; (7) \textbf{CDL-DL}: CDL with both rewards but without curriculum learning.

\begin{table*}[!htb]
    \small
    \centering
    \begin{tabular}{l|p{0.65cm}p{0.65cm}p{0.65cm}p{0.65cm}|cc|cc|cc}
    \toprule[1pt]
        \multirow{2}*{Method} & \multicolumn{4}{c|}{Embedding Metrics} & \multicolumn{2}{c|}{Diversity} & \multicolumn{2}{c|}{BLEU Scores} & \multicolumn{2}{c}{Emotion Expression} \\ \cline{2-11}
        & \hfil Avg. & \hfil Ext. & \hfil Gre. & \hfil Coh. & Dist-1 & Dist-2 & BLEU-1 & BLEU-2 & Emo-acc. & Emo-word.\\ \hline
        S2S-Attn & 0.497 & 0.352 & 0.328 & 0.582 & 0.035 & 0.119 & 0.0424 & 0.0073 & 0.244 & 0.285 \\
        EmoEmb & 0.532 & 0.381 & 0.356 & 0.594 & 0.040 & 0.133 & 0.0722 & 0.0164 & 0.693 & 0.436 \\
        EmoDS & 0.623 & 0.427 & 0.403 & 0.603 & 0.050 & 0.174 & 0.0976 & 0.0282 & 0.746 & 0.527\\
        ECM & 0.625 & 0.433 & 0.405 & 0.607 & 0.052 & 0.177 & 0.1023 & 0.0332 & 0.753 & 0.562 \\ \hline
        CDL-emo (ours) & 0.631 & 0.451 & 0.435 & 0.615 & 0.058 & 0.193 & 0.1162 & 0.0342 & 0.765 & 0.583\\
        CDL-con (ours) & 0.628 & 0.441 & 0.417 & 0.612 & 0.055 & 0.182 & 0.1059 & 0.0338 & 0.758 & 0.566 \\
        CDL-DL (ours)& 0.635 & 0.452 & 0.431 & 0.630 & 0.062 & 0.217 & 0.1187 & 0.0353& 0.794 & 0.615 \\ \hline
        CDL (ours) & \bf 0.642 & \bf 0.457 & \bf 0.438 & \bf 0.635 & \bf 0.065 & \bf 0.221 & \bf 0.1254 & \bf 0.0370 & \bf 0.823 & \bf 0.620 \\
    \bottomrule[1pt]
    \end{tabular}
    \caption{Automatic evaluation results for content and emotion measurements. The metrics Average, Extrema, Greedy, Coherence, Emotion-acc and Emotion-word are abbreviated as Avg., Ext., Gre., Coh., Emo-acc. and Emo-word., respectively.}
    \label{tab:autoresults}
\end{table*}

\begin{table*}[!htb]
    \small
    \centering
    \begin{tabular}{l|cc|cc|cc|cc|cc|cc}
    \toprule[1pt]
        \multirow{2}*{Method} & \multicolumn{2}{c|}{Like} & \multicolumn{2}{c|}{Sad} & \multicolumn{2}{c|}{Disgust} & \multicolumn{2}{c|}{Angry} & \multicolumn{2}{c|}{Happy} & \multicolumn{2}{c}{Overall} \\ \cline{2-13}
        & Con. & Emo. & Con. & Emo. & Con. & Emo. & Con. & Emo. & Con. & Emo. & Con. & Emo. \\ \hline
        S2S-Attn & 1.295 & 0.435 & 1.125 & 0.120 & 1.160 & 0.115 & \textbf{1.255} & 0.045 & 1.155 & 0.305 & 1.198 & 0.204 \\
        EmoEmb & 1.290 & 0.630 & 0.990 & 0.225 & 1.125 & 0.295 & 1.220 & 0.220 & 1.275 & 0.400 & 1.180 & 0.354 \\
        EmoDS & 1.375 & 0.685 & 1.210 & 0.395 & 1.200 & 0.340 & 1.225 & 0.345 & 1.260 & 0.535 & 1.254 & 0.460 \\
        ECM & 1.375 & 0.690 & 1.205 & 0.425 & 1.205 & 0.325 & 1.240 & 0.385 & 1.255 & 0.590 & 1.256 & 0.483 \\ \hline
        CDL & \textbf{1.395} & \textbf{0.700} & \textbf{1.245} & \textbf{0.565} & \textbf{1.235} & \textbf{0.490} & 1.250 & \textbf{0.525} & \textbf{1.305} & \textbf{0.630} & \textbf{1.286} & \textbf{0.582} \\
    \bottomrule[1pt]
    \end{tabular}
    \caption{Human evaluation results. ``Con.'' and ``Emo.'' denote content and emotion, respectively.}
    \label{tab:humanresults}
\end{table*}

\subsection{Evaluation Measures}
To better evaluate our results, we use both quantitative metrics and human judgements in our experiments.
\subsubsection{Automatic Metrics}
For automatic evaluation, we mainly choose four kinds of metrics: 1) Embedding scores (Average, Greedy, Extrema and Coherence)\footnote{We use the pre-trained word embeddings based on Sina Weibo data from \url{https://github.com/Embedding/Chinese-Word-Vectors}.} \cite{liu2016not,xu2018better}; 2) BLEU scores \cite{papineni2002bleu} in 0 to 1 scale; 3) Dist-1, Dist-2 \cite{li2016diversity} and 4) Emotion-acc, Emotion-word \cite{zhou2018emotional,song2019generating}. 

Embedding scores and BLEU scores are used to measure the quality of generated responses in terms of content relevance. Whereas, Dist-1 and Dist-2 are used to evaluate the diversity of responses\footnote{We employ a popular NLG evaluation project available at \url{https://github.com/Maluuba/nlg-eval} for automatic evaluation.}. Emotion-acc and Emotion-word are utilized to test the emotion expression. Specifically, Emo-acc is the agreement between the ground truth labels and the predicted labels through the TextCNN classifier trained before. Emo-word is the percentage of the generated responses that contain the corresponding emotion words. Since there are no multi-emotion ground truths in the test set, we only calculate the metrics between the ground truth, labelled emotion $e$, and the generated response given also label $e$ for fair comparison.

\subsubsection{Human Evaluation Settings}
Inspired by \citet{zhou2018emotional,song2019generating}, a human evaluation is conducted to better analyze the quality of generated responses. First, we randomly sample 200 queries from the test set. For each method except S2S-Attn, they generate six responses for six emotion categories, while S2S-Attn generates top 6 responses from beam search decoding for each query. Then, we send the triples of (query, response, emotion) to three human annotators without order, and require them to evaluate each response on both content level and emotion level independently. 

Content and emotion are measured by a 3-scale rating (0, 1, 2) and a 2-scale rating (0, 1), respectively. Evaluation from the content level assesses whether a response is fluent, coherent and meaningful for the query, and evaluation from the emotion level decides if a response reveals the desired emotion.

\begin{table*}[!htb]
    \footnotesize
    \centering
    \begin{tabular}{p{3.4cm}|l|l|p{3.2cm}|p{5.5cm}}
    \toprule[1pt]
        \multicolumn{1}{m{3.4cm}<{\centering}|}{\bf Query} & \multicolumn{2}{c|}{\bf Method} & \multicolumn{1}{m{3.2cm}<{\centering}|}{\bf Response} & \multicolumn{1}{m{5.5cm}<{\centering}}{\bf Translated Response} \\ \hline 
        & \multicolumn{2}{c|}{S2S-Attn} & \begin{CJK*}{UTF8}{gbsn}{生活需要改变心态。}\end{CJK*} & Life needs a change of mindset. \\ \cline{2-5}
         & \multirow{5}*{ECM} & Like & \begin{CJK*}{UTF8}{gbsn}{希望有个\textcolor[rgb]{0.7,0,0.2}{好心情}。}\end{CJK*} & Hope to have a \textcolor[rgb]{0.7,0,0.2}{good mood}. \\
        & & Sad & \begin{CJK*}{UTF8}{gbsn}{我也是很\textcolor[rgb]{0.7,0,0.2}{纠结}的。}\end{CJK*} & I am also very \textcolor[rgb]{0.7,0,0.2}{tangled}. \\
        & & Disgust &  \begin{CJK*}{UTF8}{gbsn}{你太\textcolor[rgb]{0.7,0,0.2}{过分}了。}\end{CJK*} & You are \textcolor[rgb]{0.7,0,0.2}{too much}. \\
        \multirow{1}{*}{\begin{CJK*}{UTF8}{gbsn}创意源于生活，看你对生\end{CJK*}} & & Angry & \begin{CJK*}{UTF8}{gbsn}{你怎么还没睡觉？}\end{CJK*} & how come you are still up? \\
        \multirow{1}{*}{\begin{CJK*}{UTF8}{gbsn}活的态度如何。\end{CJK*}} & & Happy & \begin{CJK*}{UTF8}{gbsn}{\textcolor[rgb]{0.7,0,0.2}{哈哈}，是啊。}\end{CJK*} & \textcolor[rgb]{0.7,0,0.2}{Haha}, yes. \\ \cline{2-5}
        Creativity comes from life, & \multirow{5}*{CDL} & Like & \begin{CJK*}{UTF8}{gbsn}{希望每天都有\textcolor[rgb]{0.7,0,0.2}{好心情}。} \end{CJK*} & Hope to have a \textcolor[rgb]{0.7,0,0.2}{good mood} every day. \\
        and depends on your & & Sad & \begin{CJK*}{UTF8}{gbsn}{我觉得我很不淡定。}\end{CJK*} & I do not think I am calm. \\
        attitude. & & Disgust & \begin{CJK*}{UTF8}{gbsn}{别说\textcolor[rgb]{0.7,0,0.2}{废话}了。}\end{CJK*} & Stop \textcolor[rgb]{0.7,0,0.2}{talking nonsense}. \\
        & & Angry &  \begin{CJK*}{UTF8}{gbsn}{你根本不懂生活！}\end{CJK*} & You do not understand life at all! \\
        & & Happy & \begin{CJK*}{UTF8}{gbsn}{\textcolor[rgb]{0.7,0,0.2}{开心}是最重要的。}\end{CJK*} &
        \textcolor[rgb]{0.7,0,0.2}{Happiness} is the most important. \\
        \midrule[1pt]
        & \multicolumn{2}{c|}{S2S-Attn} & \begin{CJK*}{UTF8}{gbsn}{我不做好事。}\end{CJK*} & I do not do good things. \\ \cline{2-5}
        & \multirow{5}*{ECM} & Like & \begin{CJK*}{UTF8}{gbsn}{是的，\textcolor[rgb]{0.7,0,0.2}{喜欢}就好。}\end{CJK*} & Yes, it is ok you \textcolor[rgb]{0.7,0,0.2}{like} it.\\
        & & Sad & \begin{CJK*}{UTF8}{gbsn}{是啊，我很\textcolor[rgb]{0.7,0,0.2}{无辜}的。}\end{CJK*} & Yeah, I am \textcolor[rgb]{0.7,0,0.2}{innocent}.\\
        & & Disgust & \begin{CJK*}{UTF8}{gbsn}{不是不是，不是好人。}\end{CJK*} & No, no, not a good person. \\
        & & Angry & \begin{CJK*}{UTF8}{gbsn}{你是什么 意思？？？}\end{CJK*} & What do you mean???\\
        \multirow{1}{*}{\begin{CJK*}{UTF8}{gbsn}善良的人还是存在的。\end{CJK*}}& & Happy & \begin{CJK*}{UTF8}{gbsn}{\textcolor[rgb]{0.7,0,0.2}{哈哈}，你也是。}\end{CJK*} & \textcolor[rgb]{0.7,0,0.2}{Haha}, you too. \\ \cline{2-5}
        Kind people still exist.& \multirow{5}*{CDL} & Like & \begin{CJK*}{UTF8}{gbsn}{你是不是在\textcolor[rgb]{0.7,0,0.2}{夸}我？}\end{CJK*} & Are you \textcolor[rgb]{0.7,0,0.2}{praising} me? \\ 
        & & Sad & \begin{CJK*}{UTF8}{gbsn}{\textcolor[rgb]{0.7,0,0.2}{可惜}我不想做。}\end{CJK*} & \textcolor[rgb]{0.7,0,0.2}{Unfortunately}, I do not want to be. \\
        & & Disgust & \begin{CJK*}{UTF8}{gbsn}{听起来像\textcolor[rgb]{0.7,0,0.2}{假话}。}\end{CJK*} & It sounds like \textcolor[rgb]{0.7,0,0.2}{a lie}. \\
        & & Angry & \begin{CJK*}{UTF8}{gbsn}{我一点也没觉得。}\end{CJK*} & I feel nothing at all.\\
        & & Happy & \begin{CJK*}{UTF8}{gbsn}{要对生活充满\textcolor[rgb]{0.7,0,0.2}{希望}。}\end{CJK*} & Be \textcolor[rgb]{0.7,0,0.2}{hopeful} about life. \\
        \midrule[1pt]
        & \multicolumn{2}{c|}{S2S-Attn} &  \begin{CJK*}{UTF8}{gbsn}{我也很喜欢。}\end{CJK*} & I like it, too. \\ \cline{2-5}
        & \multirow{5}*{ECM} & Like & \begin{CJK*}{UTF8}{gbsn}{我也\textcolor[rgb]{0.7,0,0.2}{喜欢}秋天。}\end{CJK*} & I also \textcolor[rgb]{0.7,0,0.2}{like} autumn. \\
        & & Sad & \begin{CJK*}{UTF8}{gbsn}{我也\textcolor[rgb]{0.7,0,0.2}{想念}秋天。}\end{CJK*} & I also \textcolor[rgb]{0.7,0,0.2}{miss} autumn.\\
        & & Disgust & \begin{CJK*}{UTF8}{gbsn}{你太\textcolor[rgb]{0.7,0,0.2}{过分}啦。}\end{CJK*} & You are \textcolor[rgb]{0.7,0,0.2}{too much}. \\
        \multirow{1}{*}{\begin{CJK*}{UTF8}{gbsn}我最喜欢的季节就是秋\end{CJK*}}& & Angry & \begin{CJK*}{UTF8}{gbsn}{你怎么不说我是爱？}\end{CJK*} & Why don't you say that I love it? \\
        \multirow{1}{*}{\begin{CJK*}{UTF8}{gbsn}天。\end{CJK*}}& & Happy & \begin{CJK*}{UTF8}{gbsn}{\textcolor[rgb]{0.7,0,0.2}{哈哈}，我也喜欢。}\end{CJK*} & \textcolor[rgb]{0.7,0,0.2}{Haha}, I like it too. \\ \cline{2-5}
        Autumn is my favorite & \multirow{5}*{CDL} & Like & \begin{CJK*}{UTF8}{gbsn}{秋天的天空非常\textcolor[rgb]{0.7,0,0.2}{美丽}。}\end{CJK*} & The autumn sky is very \textcolor[rgb]{0.7,0,0.2}{beautiful}. \\
        season. & & Sad & \begin{CJK*}{UTF8}{gbsn}{我很\textcolor[rgb]{0.7,0,0.2}{想念}过去的夏天。}\end{CJK*} & I \textcolor[rgb]{0.7,0,0.2}{miss} the past summer very much. \\
        & & Disgust & \begin{CJK*}{UTF8}{gbsn}{秋天的景色很\textcolor[rgb]{0.7,0,0.2}{难看}。}\end{CJK*} & The autumn scenery is \textcolor[rgb]{0.7,0,0.2}{ugly}. \\
        & & Angry & \begin{CJK*}{UTF8}{gbsn}{你\textcolor[rgb]{0.7,0,0.2}{有病}吧？}\end{CJK*} & Are you \textcolor[rgb]{0.7,0,0.2}{insane}? \\
        & & Happy & \begin{CJK*}{UTF8}{gbsn}{\textcolor[rgb]{0.7,0,0.2}{哈哈}，要不要去秋游？}\end{CJK*} & \textcolor[rgb]{0.7,0,0.2}{Haha}, do you want to go to the autumn tour? \\ 
    \bottomrule[1pt]
    \end{tabular}
    \caption{Sample responses generated by S2S-Attn, ECM and CDL (original Chinese and English translation). The colored words are the emotion words corresponding to the given emotion category.}
    \label{tab:case_study}
\end{table*}

\subsection{Experimental Results}
Now we demonstrate our experimental results on both automatic evaluation and human evaluation.

\subsubsection{Automatic Evaluation Results}
The automatic results are shown in Table \ref{tab:autoresults}. The top part is the results of all baseline models, and we can see that CDL outperforms the other methods on all metrics (t-test, $p$-value $<$ 0.05). The improvements of CDL on \textit{Coherence}, \textit{Emotion-acc} and \textit{Emotion-word} are significant, indicating that it can enhance content consistency and emotion expression simultaneously. EmoDS and ECM have similar performance, as both of them use the forward method to pay more attention on the emotion factor. S2S-Attn can only generate fluent responses based on semantic mapping, but fail to express diverse responses.

The bottom part of Table \ref{tab:autoresults} shows the results of our ablation study. Comparisons among CDL-emo, CDL-con and CDL show the effectiveness of the combined reward for both emotion expression and content consistency. In addition, we can find that with the support of curriculum learning, CDL can achieve better results than CDL-DL. 

\subsubsection{Human Evaluation Results}
The results are shown in Table \ref{tab:humanresults}. CDL obtains the best performance (t-test, $p$-value $<$ 0.05) on both emotion expression (0.582) and content coherence (1.286). As we can see, there is no obvious difference between EmoDS and ECM. Due to the insufficient training data of ``Anger'' (79,611 in queries and 138,198 in responses), S2S-Attn achieves the best content score for it, which is similar to the results of \citet{zhou2018emotional}.

\begin{table}[!htb]
    \small
    \centering
    \begin{tabular}{l|cccccc}
    \toprule[1pt]
        Method (\%) & 2-1 & 1-1 & 0-1 & 2-0 & 1-0 & 0-0 \\ \hline
        S2S-Attn & 10.3 & 7.2 & 2.8 & 36.4 & 26.5 & 16.8 \\
        EmoEmb & 21.8 & 12.6 & 7.5 & 24.6 & 15.3 & 18.2 \\
        EmoDS & 28.7 & 15.6 & 4.0 & 22.7 & 13.5 & 15.5 \\
        ECM & 27.1 & 12.7 & 4.5 & 23.5 & 15.4 & 16.8 \\ \hline
        CDL & \textbf{32.5} & \textbf{17.6} & 4.1 & 17.7 & 12.8 & 15.3 \\
    \bottomrule[1pt]
    \end{tabular}
    \caption{The percentage of responses in human evaluation of \textit{Content-Emotion} scores. 2-1 means content score is 2 and emotion score is 1.}
    \label{tab:distribution}
\end{table}

Results of emotion and content in Table \ref{tab:humanresults} are independent. To better evaluate the overall quality of the generated responses, we present results in Table \ref{tab:distribution} by considering content and emotion scores simultaneously. 32.5\% of the responses generated by CDL are annotated with Emotion score 2 and Content score 1, which shows that CDL is better at producing coherent as well as emotion-rich responses.

Agreements to measure the consistency among three annotators are calculated with the Fleiss' kappa \cite{fleiss1973equivalence}. Fleiss' kappa for content and emotion is 0.497 and 0.825, indicating ``Moderate agreement'' and ``Substantial agreement'', respectively. 

\subsection{Case Study}
Table \ref{tab:case_study} shows the examples generated by S2S-Attn, ECM and CDL. As can be seen from it, for a given post, there are multiple emotion categories that are appropriate for its response in the conversation. S2S-Attn generates a response with a random emotion, while ECM and CDL can utilize the specific emotion label. Compared with ECM, CDL can generate both coherent and informative responses with any desired emotion. In addition, the emotion can be expressed in either explicit or implicit manner. For example, ``\begin{CJK*}{UTF8}{gbsn}{你/根本/不懂/生活！}\end{CJK*} (You do not understand life at all!)'' express anger when we read this sentence as a whole, while ``\begin{CJK*}{UTF8}{gbsn}{美丽}\end{CJK*} (beautiful)'' or ``\begin{CJK*}{UTF8}{gbsn}{开心}\end{CJK*} (happy)'' are strong emotion words to represent ``Like'' or ``Happy''.

\subsection{Further Analysis of CDL}
\begin{figure}[htb]
\begin{center}
   \includegraphics[width=1.0\linewidth]{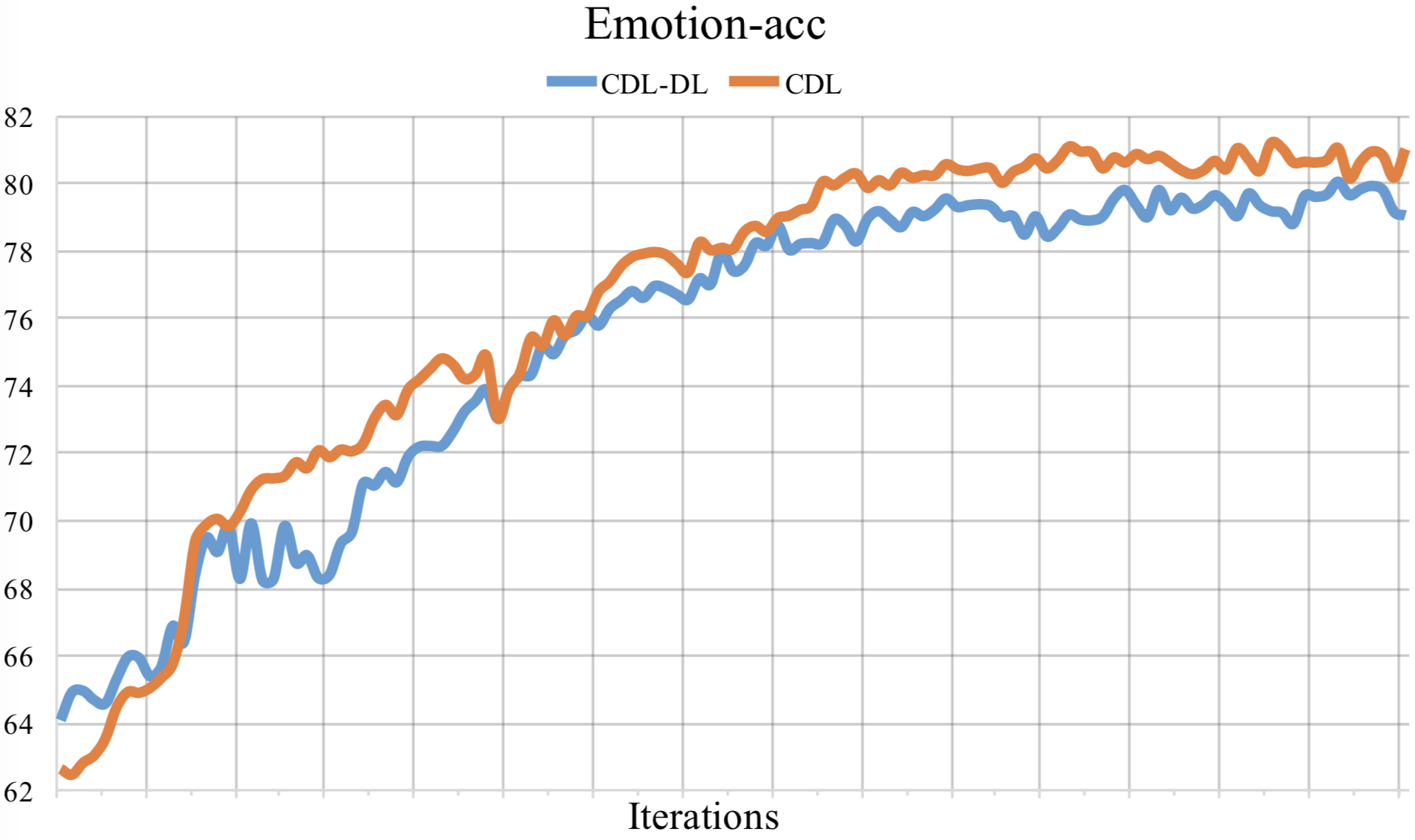}
\end{center}
   \caption{Comparison of CDL and CDL-DL for \textit{Emotion-acc} on the validation set.}
\label{fig:acc_change}
\end{figure}

Here, we conduct a further analysis to show some characteristics of this task and the effect of CDL. Emotion lexicon size and classification accuracy after pre-training of each category ($N$(correct prediction) $\div$ category size) are listed in Table \ref{tab:analysis_statistics}. We can see that the classification accuracy is not totally related to the emotion lexicon size, indicating the emotion expression is partially implicit or explicit. To better illustrate the learning efficiency of CDL, we plot the changes of \textit{Emotion-acc} on the validation set. As shown in Figure \ref{fig:acc_change}, CDL accelerates the learning effectively and consistently outperforms CDL-DL.

\begin{table}[!htb]
    \small
    \centering
    \begin{tabular}{p{1.2cm}|ccp{0.8cm}cc}
    \toprule[1pt]
         & Like & Sad & Disgust & Angry & Happy \\ \hline
         Lex. Size & 1,629 & 294 & \hfil 1,142 & 30 & 405\\ \hline
         ACC ($f$) & 0.653 & 0.691 & \hfil 0.609 & 0.736 & 0.818\\ \hline
         ACC ($b$) & 0.690 & 0.655 & \hfil 0.602 & 0.756 & 0.808 \\
    \bottomrule[1pt]
    \end{tabular}
    \caption{Emotion lexicon size and classification accuracy after pre-training of each emotion category. ``Lex.'', ``ACC($f$)'' and ``ACC($b$)'' represent lexicon, classification accuracy of forward process and classification accuracy of backward process, respectively.}
    \label{tab:analysis_statistics}
\end{table}

\section{Related Work}
\label{sec:relatedwork}
Responses generated by traditional open-domain dialogue systems are usually safe and generic. To produce diverse and informative responses, researchers tried to either import latent variables for model construction \cite{zhao2017learning,serban2017hierarchical,shen2019modeling} or utilize some extra knowledge, e.g., sentence types, personas, emotions, documents and knowledge triples/graphs \cite{ke2018generating,li2016persona,zhou2018emotional,meng2019refnet,zhou2018commonsense,niu2019knowledge}. In this paper, we mainly touch on two branches of research: emotional response generation and dual learning in NLP.

\subsection{Emotional Response Generation}
\label{sec:related_1}
Early studies have proven that dialogue systems with proper emotional expressions and reactions can directly improve user satisfaction \cite{prendinger2005empathic,prendinger2005using} and contribute to effective users' performance \cite{partala2004effects}. \citet{polzin2000emotion} and \citet{polzin2000emotion} apply rule-based methods to choose emotional responses from a conversation corpus, but those rules are hard to extend to large corpora. 
With the advent of deep learning, some researchers utilize neural networks to solve this problem \cite{ghosh2017affect,hu2017toward,zhou2018mojitalk,sun2018emotional}. 
Besides, the Valence, Arousal, and Dominance (VAD) lexicon \cite{warriner2013norms,mohammad2018obtaining} is embedded to the sequence-to-sequence model \cite{sutskever2014sequence} to provide extra affective information \cite{asghar2018affective,zhong2019affect}.

Responses generated by above studies can simply continues the emotion of the query. To generate emotion-controllable responses, \citet{zhou2018emotional} address the emotion factor in large-scale conversations, and propose ECM to generate responses based on different given emotions. After that, \citet{colombo2019affect} augment ECM with VAD embeddings and modified the loss function and decoding procedure. \citet{song2019generating} use lexicon-based attention and a word-based classifier to improve the ability of emotion expression.

\subsection{Dual Learning in NLP}
\label{sec:related_2}
\citet{he2016dual} propose Dual Learning (DL) for machine translation first which consider the source to target language translation and target to source language translation as a dual task. After that, \citet{tang2017question} implement a dual framework for the question answering system. Both \citet{zhang2018reinforcing} and \citet{cui2019dal} use similar idea in dialogue generation task to produce coherent but not safe responses, since they find that a more diverse and specific response usually has a higher probability of being transformed back to the given query. \citet{luo2019dual} and \citet{luo2019towards} exploit DL in unsupervised text style transfer to relieve the need of parallel data.

The differences between our method and those in Section \ref{sec:related_1} and Section \ref{sec:related_2} are: (1) We consider the emotion expression and content consistency simultaneously via a DL method.
(2) Instead of regarding the query as an emotionless sentence, we utilize the emotion of query, which can help model the emotion shifting and coherence to improve the quality of response. (3) To better model the changes in emotion and content between the query and response, we combine the DL method with curriculum learning, which is known to improve the effectiveness and generalization.

\section{Conclusion}
\label{sec:conclusion}
In this paper, we propose a new framework Curriculum Dual Learning (CDL) for generating emotional responses in a controlled manner. Since existing methods in this field only focus on the emotion expression of target label but fail to consider the emotion of queries, the safe response problem deteriorates and hurts the content consistency. CDL utilizes two kinds of rewards to enhance emotion and content simultaneously via dual learning. Besides, with the support of curriculum learning, it can be more efficient. Experimental results show that CDL can generate fluent, coherent, informative as well as emotional responses.

\section*{Acknowledgements}
This work was supported by National Key R\&D Program of China (NO. 2017YFE0192900). We sincerely thank the anonymous reviewers for their helpful and valuable suggestions.

\bibliography{anthology,acl2020}
\bibliographystyle{acl_natbib}
\end{document}